\documentclass{article}
\usepackage{spconf,amsmath,graphicx}
\usepackage{caption}

\usepackage{color}
\usepackage[normalem]{ulem}

\def\nonlinearities{
    \begin{bmatrix}
        \text{sigm} \\
        \text{sigm} \\
        \text{sigm} \\
        \text{tanh} \\
    \end{bmatrix}}

\title{Understanding Recurrent Neural State Using Memory Signatures}
\twoauthors
  {Skanda Koppula\sthanks{This author completed this work
	while working at Google, Inc.}}
	{MIT}
  {Khe Chai Sim, Kean Chin}
	{Google, Inc.}
\begin{document}
\ninept
\maketitle
\begin{abstract}
We demonstrate a network visualization technique to analyze the recurrent state inside the LSTMs/GRUs used commonly in language and acoustic models. Interpreting intermediate state and network activations inside end-to-end models remains an open challenge. Our method allows users to understand exactly how much and what history is encoded inside recurrent state in grapheme sequence models. Our procedure trains multiple decoders that predict prior input history. Compiling results from these decoders, a user can obtain a signature of the recurrent kernel that characterizes its memory behavior. We demonstrate this method's usefulness in revealing information divergence in the bases of recurrent factorized kernels, visualizing the character-level differences between the memory of n-gram and recurrent language models, and extracting knowledge of history encoded in the layers of grapheme-based end-to-end ASR networks.
\end{abstract}
\section{Introduction}
\label{sec:intro}

Recurrent neural networks such as LSTMs and GRUs achieve state-of-art performance in a variety of sequence-based tasks, such as language modeling \cite{lstmlm}, acoustic modeling \cite{lstmam}, end-to-end ASR \cite{las}, and machine translation \cite{translation}. Interpreting the parameters and output of pre-trained recurrent networks, however, is difficult. Traditional statistics-based models, such as n-grams or regression, allow for convenient interpretation with respect to their features and training data. Deep neural networks, however, learn opaque transforms and embeddings in different real-valued, continuous spaces across the depth of the network. For the purpose of understanding the behavior and limitations of a network, it is sometimes useful to be able to characterize the information contained in a network's parameters.

In this paper, we explore the use of state decoders to extract information embedded inside the recurrent state produced by pre-trained language and speech networks. We employ this to characterize and dissect the recollection ability of LSTM kernels. As we will demonstrate, such information can help us answer questions about the behavior of cross-domain model adaptation methods (such as model retraining and kernel factorization), and information encoded within the layers of an end-to-end ASR network such as Listen-Attend-Spell \cite{retraining, factorized, las}.

The following section describes the prior art in network visualization and interpretation, and where our work stand in relation. Section \ref{sec:architecture} describes our visualization technique and our base recurrent architecture. We describe experiments applying of our method in Sections \ref{sec:overall}, \ref{sec:interpolation}, \ref{sec:factorized}, and \ref{sec:las}. We conclude with a brief discussion of our results in Section \ref{sec:conclusion}.

\section{Prior Work}
Prior methods used to visualize and interpret deep neural networks fall into three main classes:
\begin{enumerate}
\item Methods that regularize the activations or alter the structure of the network during training time to align with interpretability metrics \cite{stimulatedlearning, hmmrnn, affinenetworks, khechaiinterp2}.
\item Methods that operate on pre-trained models to identify inputs or activations that strongly contribute to the network's output \cite{patternnet, activationmaximization, deconvolution, pertubingweights, deeplift, extra1, extra2}.
\item Methods such as DeepDream or image inversion that attempts to understand properties of the filter weights that define model behavior \cite{icasspviz, inversion}.
\end{enumerate}

Our method falls into this last category, extending this class to include visualization of the memory capacity of time-dependent recurrent kernels. Prior work on interpreting LSTMs in context of language or speech has been focused on identifying important input tokens in sentiment classification \cite{nlplstm}, correlating LSTM inputs with language model outputs \cite{karpathy}), or converting language model LSTMs to rule-based classifiers \cite{ruleextraction}. Our visualization method introduces a way to understand the information encoded inside RNN state, in grapheme-based language and end-to-end ASR models.

\section{Preliminaries}
\label{sec:preliminaries}

In the stacked LSTM formulation \cite{lstm, karpathy}, the internal recurrent state $\mathbf{c}_t^l$ and cell output $\mathbf{h}_t^l$ at time-step $t$ and layer $l$ are given by forward update:
\begin{equation}
    \mathbf{c}_t^l = \mathbf{f} \cdot \mathbf{c}_{t-1}^l + \mathbf{i} \cdot \mathbf{I}
\quad \quad
\mathbf{h}_t^l = \mathbf{o} \cdot \text{tanh}(\mathbf{c}_t^l)
\end{equation}
where $(\mathbf{f,i,o,I})$ are the forget-gate, input-gate, output-gate, and projected-input vectors, respectively. These vectors are computed using a $[4n \times 2n]$ kernel weight matrix $\mathbf{W}_l$:
$$
[\mathbf{f,i,o,I}] = \nonlinearities (\mathbf{W}_l \cdot \binom{\mathbf{h}_t^{l-1}}{\mathbf{h}_{t-1}^l})
$$
Gated Recurrent Units (GRUs) are a simpler cell formulation that combines the recurrent state and output into one vector, $\mathbf{h}_t^l$ \cite{grus}. Output $\mathbf{h}_t^l$ is the smooth interpolation between a candidate output $\mathbf{\widetilde{h}}_t^l$ and the previous cell output $\mathbf{h}_{t-1}^l$. For brevity, we leave interested readers to refer to \cite{karpathy} for a rigorous exposition of both kernel cell types.

\section{Extracting Memory Signatures}
\label{sec:architecture}

Of interest in recurrent networks is the information contained in $\bf{c}_{t-1}$ and $\bf{h}_{t-1}$ about the history of prior inputs. For example, the fact that $\mathbf{W}_l$ encodes a memory behavior such that the cell forgets all inputs prior to six time steps ago could be indicative of a domain-specific pattern. In another example, the fact that the kernel's memory vectors ($\bf{c}_{t}$ or $\bf{h}_{t}$) retain information about the character $t$ but not $r$ reflects asymmetric structure in the underlying set of modeled sequences. Importantly, complete or partial input recall from $\bf{c}_t$ and $\bf{h}_t$ is the first step in maintaining long-term dependencies.

\begin{figure}[t]
  \begin{minipage}[b]{1.0\linewidth}
  \centering
  \centerline{\includegraphics[width=8.5cm]{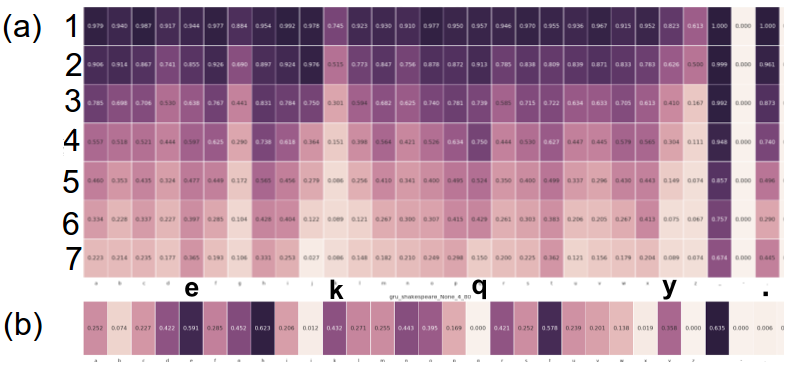}}
  \captionof{figure}{\textbf{a.} Memory signature for a 1-layer GRU trained on the Shakespeare corpus. $y$-axis is time-steps in the past for back-prediction. $x$-axis is the list of symbols \texttt{[a-z ,.]}. \textbf{b.} Character-level accuracy during forward-prediction. In both figures, accuracy ranges from dark purple (100\% recall) to white (0\% recall).}\medskip
  \label{fig:confusion}
  \end{minipage}
\end{figure}

To create our learned visualization construction, the \textit{memory signature} of a cell, we train a set of $\Delta \times L$ decoders to extract information from $\mathbf{c}_t^l$. Every decoder $(\delta, l)$ is responsible for decoding the states in layer $l$ and recalling input $\delta$ time steps prior, $1 \leq \delta \leq \Delta$. The $(\delta=2,l=1)$-decoder, for example, learns to predict the discrete input at time-step $t-2$ given $\mathbf{c}^1_t$. During training of the decoders, the weights of the recurrent network are fixed. In a well-tuned decoder, the accuracy and confusion matrix of the decoder is indicative of the ability of the kernel's ability to remember discrete inputs from $t$-time steps ago. The memory signature is a compilation of the accuracies of the decoder, either the accuracy across discrete inputs (e.g. Figure~\ref{fig:confusion}) or the overall accuracy for each time step (Figure~\ref{fig:statesize}).
The decoders are trained in a similar fashion to the primary model, using the corpus's training partition, and tested using the evaluation partition.

In sections \ref{sec:overall}, \ref{sec:interpolation}, and \ref{sec:factorized}, our experiments use character-level recurrent networks. These networks follow the same structure as in \cite{karpathy}: 
$$
\widetilde{g}_{t+1} = \text{argmax}\ \mathbf{W}_P(RNN(\mathbf{W}_C \cdot \text{one-hot}(g_t))) + \mathbf{b}_P)
$$
where $g_t$ is the grapheme input, $\widetilde{g}_{t+1}$ is the predicted next grapheme, $\mathbf{W}_C$ is a character embedding, and $\mathbf{W}_P,\mathbf{b}_P$ is a projection from the RNN dimension to the dimension of the symbol set. We use fully-connected networks with ReLU non-linearities for decoding. Language model perplexities and decode accuracies are reported with optimal training dropout on the decoder layers and recurrent state. These were found by sweeping  dropout keep probabilities from [0.5, 1] in intervals of 0.1. Our decoder parameter count generally settled at being at least as large as the combined kernel and input projection parameter count. In our experiments in sections \ref{sec:overall}, \ref{sec:interpolation}, and \ref{sec:factorized}, this setting resulted in best decoder performance.

In the experiments in subsequent sections, we will use two datasets: the complete works of Shakespeare (with 83k text segments, as used in \cite{hmmrnn}) and the Wall Street Journal (WSJ) corpus with 37K text segments (\texttt{SI-284} for training, \texttt{dev93} for development and early stopping, and \texttt{eval92} for evaluation) \cite{wsj}. Unless otherwise noted, our grapheme models are trained over the symbol set \texttt{[a-z .,\#']}, with \# replacing any digit \texttt{[0-9]}.

Figure~\ref{fig:confusion} illustrates input recall and forward prediction of a GRU trained on the Shakespeare corpus. We see a continuous decline in recall rates as history length increases, different between characters, as expected. Interestingly, a model's ability to remember a character in its recurrent state correlates poorly with its prediction accuracy. This is noted in some of the specific characters highlighted in the figure: the model has relatively poor accuracy predicted the instances of \texttt{q} in a sequence, but keeps a comparatively strong memory of `\texttt{q}'. The inverse trend is noted for `\texttt{y}'. The model predicts future occurences of `\texttt{y}' with high accuracy (presumably because it is well represented in the training prior), but finds that it does not need to keep strong memory of `\texttt{y}'. Symbols `\texttt{k}', `\texttt{e}' and `\texttt{.}' exhibit a similar pattern. The divergence between character accuracy and character memory is a trend noticed in other models in our experiments.

\section{Kernel Size and Depth on Recall}
\label{sec:overall}
\begin{figure}[t]
\begin{minipage}[b]{.48\linewidth}
  \centering
  \centerline{\includegraphics[width=4.0cm]{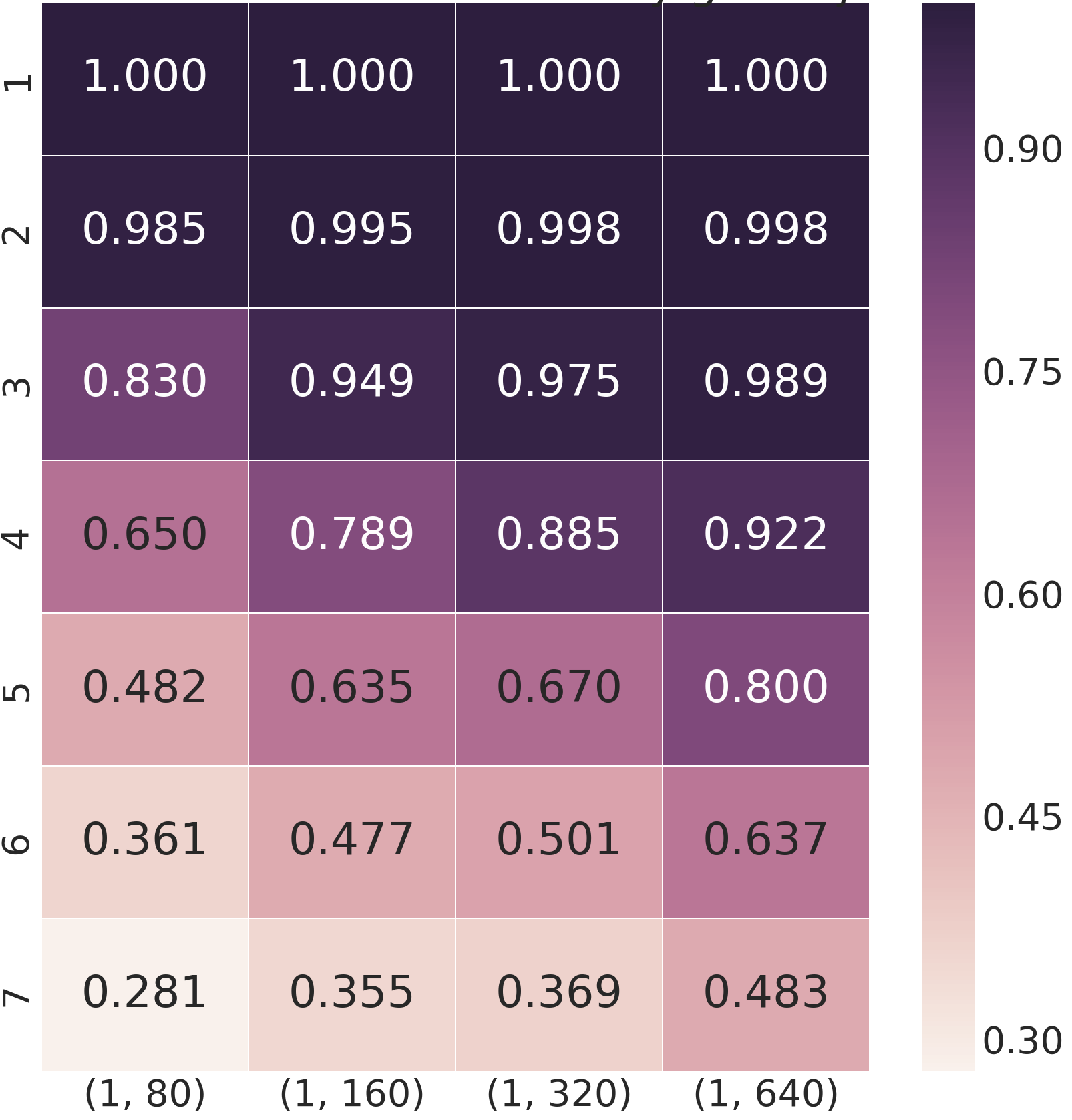}}
  \centerline{(a) GRU}\medskip
\end{minipage}
\hfill
\begin{minipage}[b]{0.48\linewidth}
  \centering
  \centerline{\includegraphics[width=4.0cm]{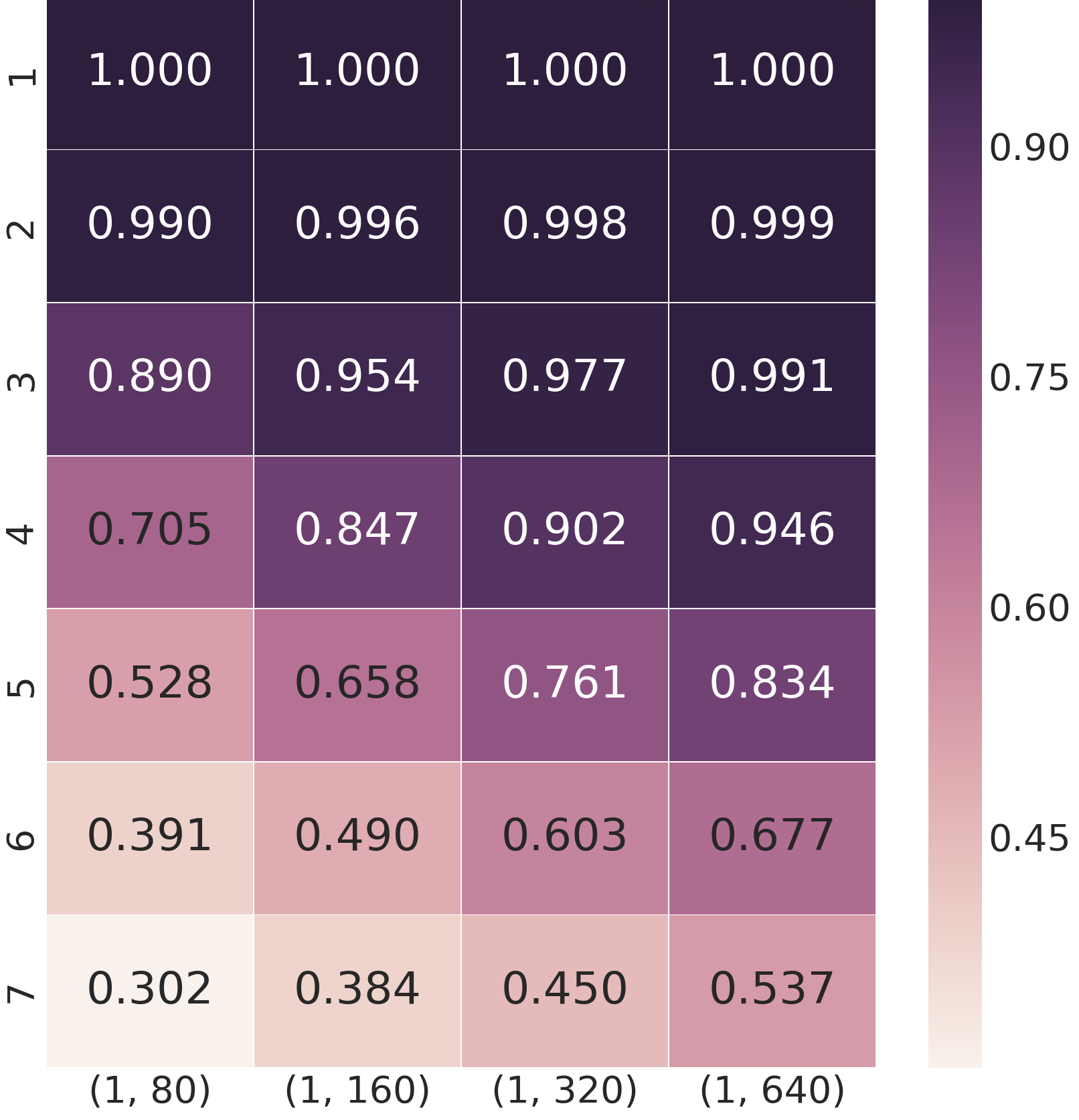}}
  \centerline{(b) LSTM}\medskip
\end{minipage}
\vspace{-0.1cm}
\caption{Marginal recall accuracies of a 1-layer GRU and LSTM language models, trained on WSJ, with varying kernel dimensionalities. $y$-axis is the number of time-steps in the past being predicted (same as in Figure~\ref{fig:confusion}). $x$-axis describes the model architecture in format \texttt{(number of layers, kernel size)}; from left to right, [1,80], [1,160], [1,320], [1,640].}
\label{fig:statesize}
\end{figure}

In our first set of experiments, we investigate the effect that state size and network depth have on the encoded memory in GRU and LSTM-based language models. Figure~\ref{fig:statesize} illustrates the effect of increasing state size on input recall. Moving from state size 80 to 640 not only continuously decreased the test perplexity of both language models by 15\%, but also significantly increased the amount of information encoded in the recurrent state about prior input. This is expected, as larger state has greater capacity to store information. We experimented with a variety of decoder sizes and dropout percentages, selecting the best decoder architecture for every model architecture. We notice that the GRU exhibits slightly worse recall from memory, possibly because of the smaller parameter count for the corresponding state size.

\begin{figure}[t]
\begin{minipage}[b]{.48\linewidth}
  \centering
  \centerline{\includegraphics[width=4.0cm]{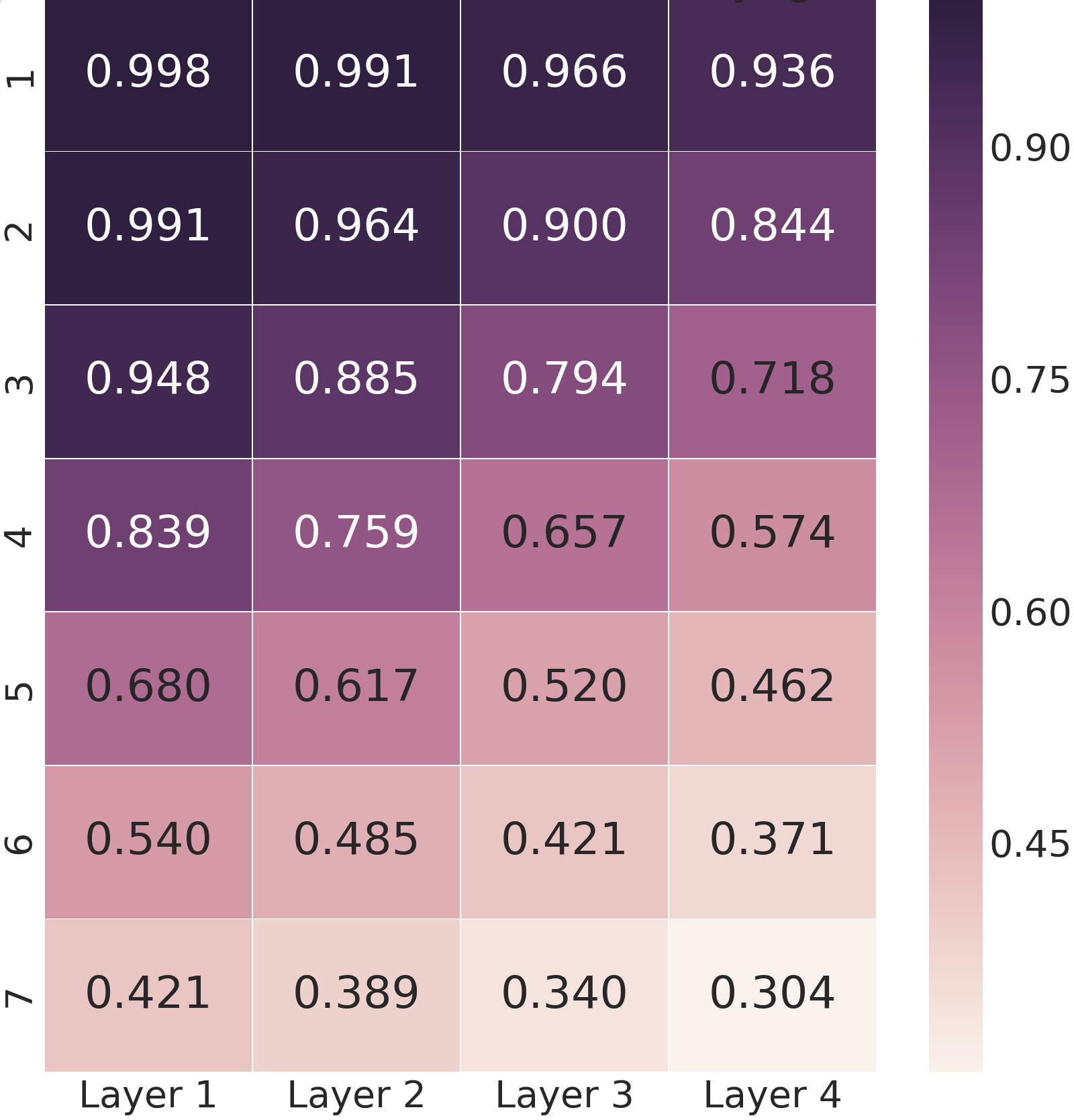}}
  \centerline{(a) GRU}\medskip
\end{minipage}
\hfill
\begin{minipage}[b]{0.48\linewidth}
  \centering
  \centerline{\includegraphics[width=4.0cm]{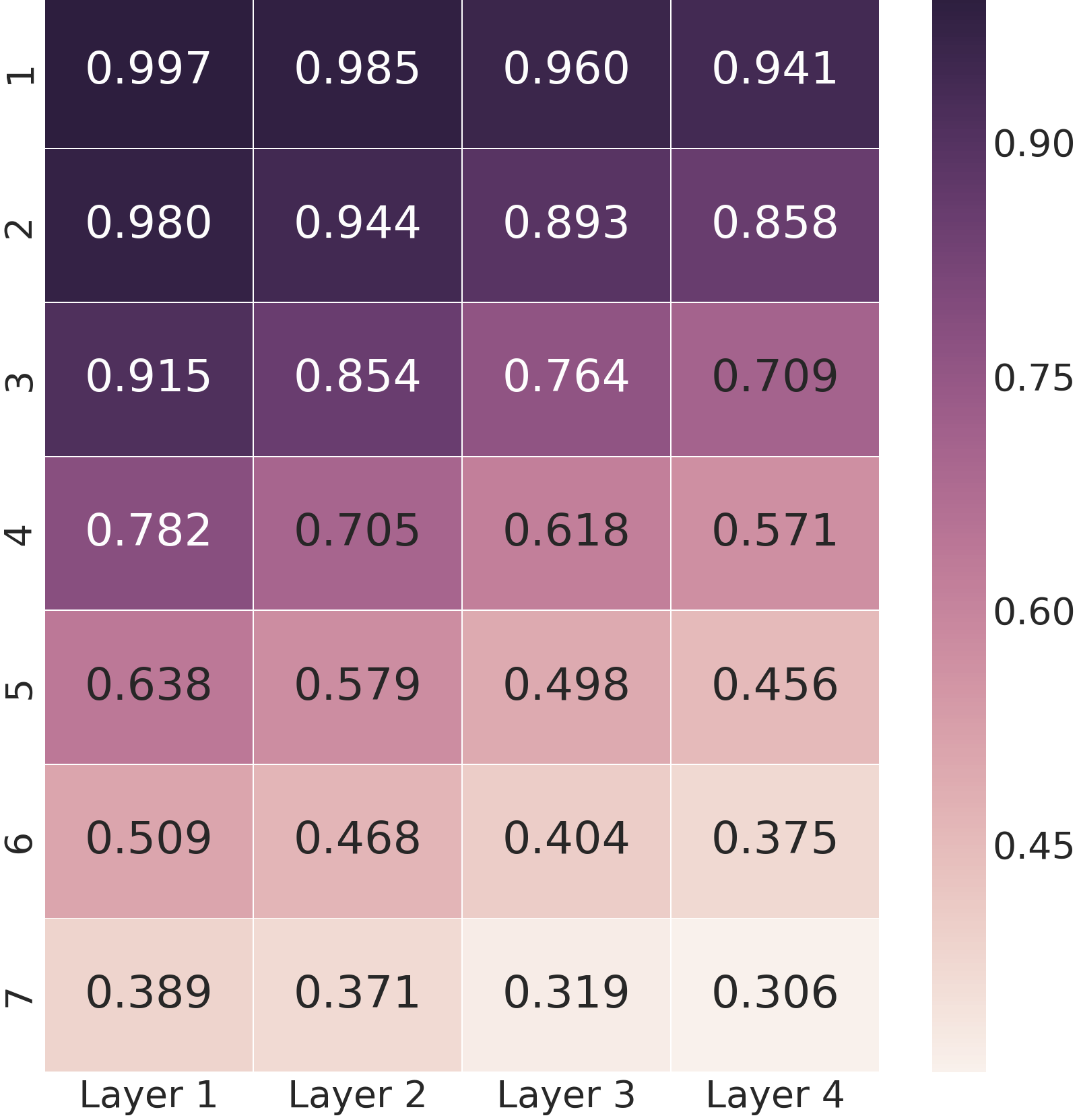}}
  \centerline{(b) LSTM}\medskip
\end{minipage}
\caption{Marginal recall accuracies for the recurrent state within each layer of a \texttt{[4,320]}-sized GRU and LSTM WSJ LM. The $y$-axis is the same as in Figure~\ref{fig:statesize}. The $x$-axis labels the recurrent layer from which state is extracted: Layer 1 (closest to input), Layer 2, Layer 3, and Layer 4.}
\label{fig:statedepth}
\end{figure}

Figure~\ref{fig:statedepth} illustrates the amount of decodable history maintained in the recurrent state of each layer as a function of depth of the network. We see the trend that decodable knowledge of prior input decreases as information propogates through layers of the recurrent network (left to right). This is likely because (1) the information content becomes strongly encoded as it passes through more transforms and (2) memory of input is summarized and filtered out at every layer, as deemed fit by a trained network. In an effort to increase the decoder strength, to investigate possibility (1), we doubled the decoder depth on layer 3 and 4, and saw small improvements to the recall accuracy ($<5$\%). This is an avenue for future investigation.

\section{Domain Adaptation via Re-training}
\label{sec:interpolation}

An open problem in language modeling is cross-domain adaptation: given a large amount of data from a non-target domain, and a small amount (or no) data from a target domain, how do we create the best langauge model for the target domain? One common approach for n-gram LMs is interpolation between an n-gram derived from the out-of-domain data and an n-gram derived from in-domain data \cite{crossdomain}. In the recurrent networks world, a commonly applied technique for such domain adaptation is to train an LSTM on the out-of-domain data, and subsequently re-train the network on subset of available in-domain data. In the process of retraining, what happens to the character recall behaviors?

\begin{figure}[t]
  \begin{minipage}[b]{1.0\linewidth}
  \centering
  \centerline{\includegraphics[width=8.5cm]{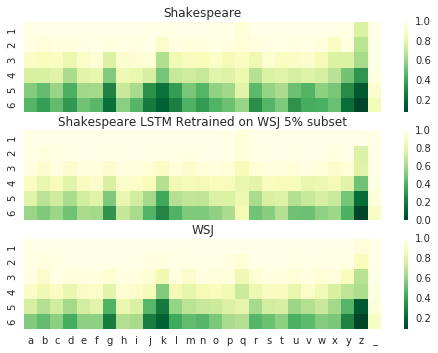}}
      \captionof{figure}{Memory signatures for a 1-layer LSTM trained on the \textbf{a.} Shakespeare corpus \textbf{b.} Shakespeare corpus, retrained on a WSJ 5\% dataset \textbf{c.} WSJ corpus. Axis are the same as in Figure~\ref{fig:confusion}.a. \textit{Lighter} green represents \textit{higher} accuracy of input recall.}\medskip
  \label{fig:retraining}
  \end{minipage}
\end{figure}

The out-of-domain LSTM, trained on Shakespeare and evaluated on WSJ, has test perplexity of 6.791 and memory signature as shown in Figure~\ref{fig:retraining}.a. Re-training our Shakespeare LSTM with 5\% of WSJ (less than 2\% of the out-of-domain data), we obtain the signature in Figure~\ref{fig:retraining}.b. We see qualitatively that signature Figure~\ref{fig:retraining}.b. more strongly resembles the signature of the WSJ-trained LSTM (Figure~\ref{fig:retraining}.c.) than a Shakespeare trained LSTM: the weak `\texttt{k}',`\texttt{l}',`\texttt{m}' symbol memory, the weaker `\texttt{b}',`\texttt{c}' memories, and weaker `\texttt{u}' memory. Accordingly, the WSJ-eval perplexity difference between the retrained model and pure WSJ, 4.110 - 3.443 = 0.667, is 4x smaller than the corresponding difference between the retrained model and a pure Shakespeare trained model, 6.791 - 4.110 = 2.681. In these models, and other retrained models in our experiments, memory signatures act as a strong visual indicator for model similarity.

\begin{figure}[t]
  \begin{minipage}[b]{1.0\linewidth}
  \centering
  \centerline{\includegraphics[width=8.5cm]{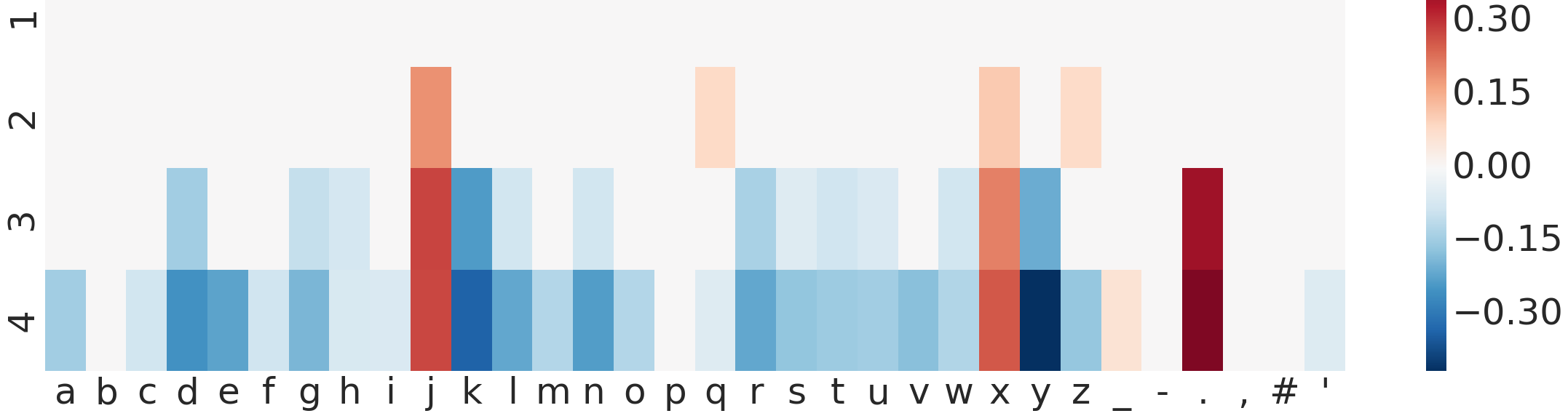}}
      \captionof{figure}{Difference in accuracy of character recall between a 1-layer GRU and 5-gram trained on WSJ. Axis are the same as in Figure~\ref{fig:retraining}. Red cells represent stronger GRU recall, blue cells represents stronger n-gram recall.}\medskip
  \label{fig:difference}
  \end{minipage}
\end{figure}

As a related aside, using memory signatures we can visualize an aspect of complementarity between n-gram and LSTM language models. Work by Figure~\ref{fig:difference} plots the differences between the input recall capacity of a 5-gram (derived from states encoded in the n-gram) and a 1-layer GRU trained on WSJ. The strong character-level memory differences between the two grapheme model types corroborate the empirical perplexity advantages seen combining n-gram and recurrent networks in \cite{ngramlstmcombo}, and work done to use n-gram posteriors to bootstrap LSTM language model training \cite{ngramlstmcombo2}.

\section{Information Diversity in Factorized Recurrent Kernels}
\label{sec:factorized}

A recent technique applied for domain adaptation of recurrent networks is weight matrix decomposition by \cite{factorized, factorized2}. In brief, every recurrent kernel $W_l$, is decomposed as the sum of a primary weight matrix and secondary weight bases, weighted by a $\mathbf{\lambda}$ vector:
$$
\bf{W_l} = W_0 + \lambda_0 W_0 + \lambda_1 W_1 + \ldots + \lambda_n W_n
$$
$\mathbf{\lambda}$ represents statistics of the input domain. For example, in our experiments, we feed normalized bigram frequencies as $\mathbf{\lambda}$, and correspondingly use $34^2 = 1156$ rank-one bases. While \cite{factorized} has shown empirical advantages to this network structure for generalizing across domains, it remains to be answered whether each of the kernel bases contains divergent information from each other, or whether there are basis after training that contain little information.

\begin{figure}[t]
  \begin{minipage}[b]{1.0\linewidth}
  \centering
  \centerline{\includegraphics[width=8.5cm]{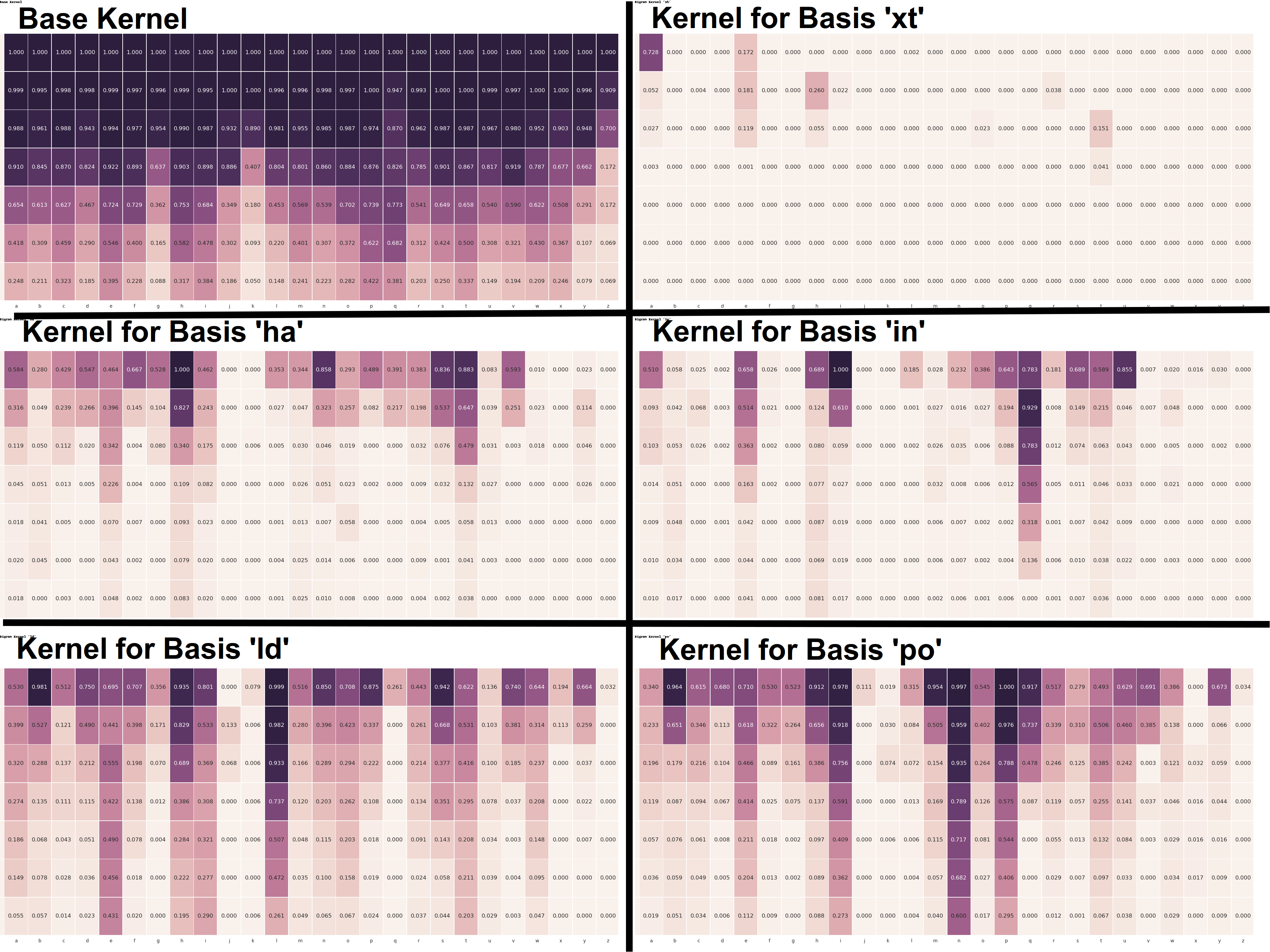}}
      \captionof{figure}{Memory signatures (in the same format as Figure~\ref{fig:confusion}) of the adaptation base kernel, and bi-grams `\texttt{xt}', `\texttt{ha}', `\texttt{in}', `\texttt{ld}', and `\texttt{po}' (top-left to bottom-right). Accuracy ranges from dark purple (100\% recall) to white (0\% recall).}\medskip
  \label{fig:bases}
  \end{minipage}
\end{figure}

We can take a look at this question from the perspective of kernel memory: do the individual basis encode different memory signatures, implying the have different responisibilities in maintaining long term dependencies? Our experiments suggest the answer is yes, as shown in Figure~\ref{fig:bases}.

The base kernel dominates model's ability to maintain memory. The other five randomly-selected base kernels learn quite different memory behaviors. In fact, we see interesting correlation between characters recollected and the corresponding $\lambda$-bigram for that kernel; for example, the `\texttt{he}' bi-gram kernel recalls `\texttt{t}' and `\texttt{h}' well. Our memory signatures suggest that models with this factorized structure learn non-overlapping knowledge. We have the capability now, for example, to prune the kernels which contain memory of graphemes that are not in our target domain.

\section{End-to-End ASR with Listen-Attend-Spell}
\label{sec:las}

\begin{figure}[t]
  \begin{minipage}[b]{1.0\linewidth}
  \centering
  \centerline{\includegraphics[width=8.5cm]{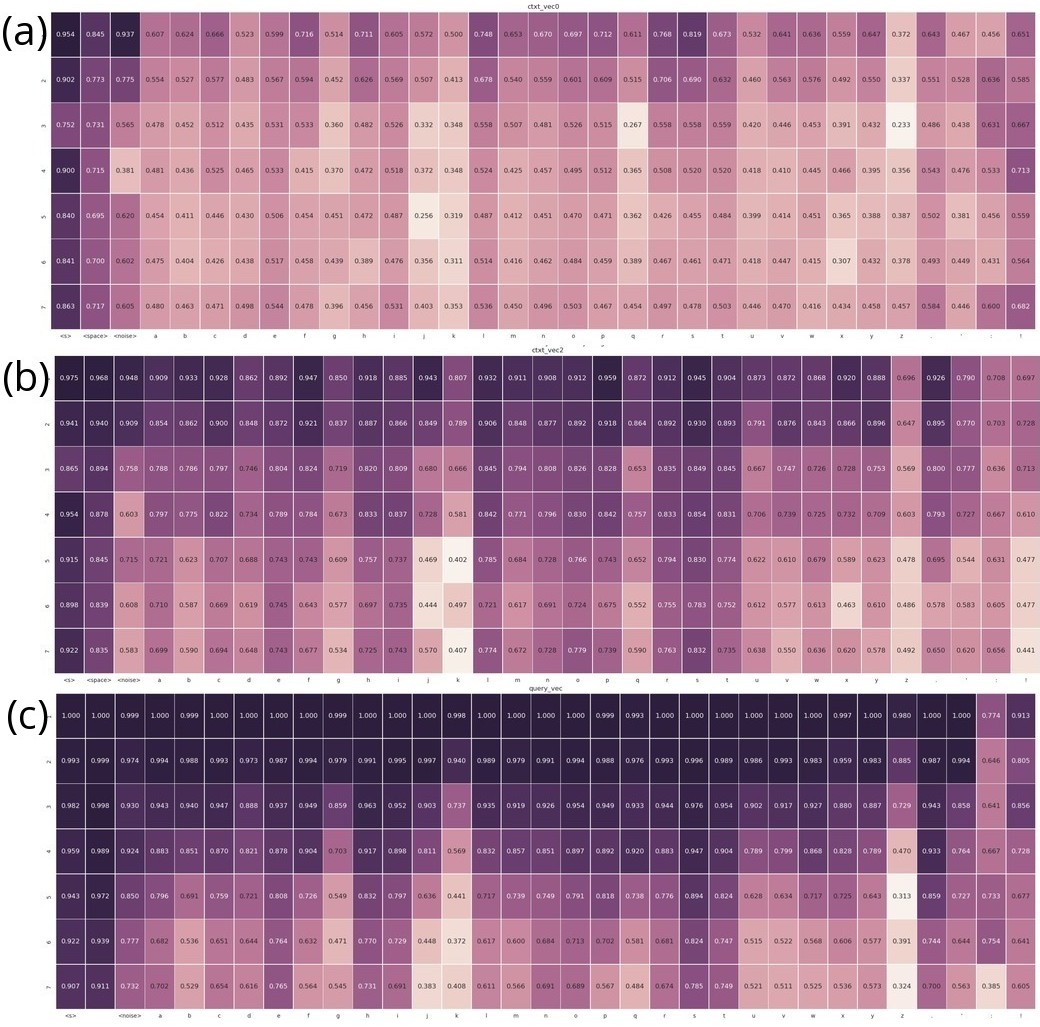}}
  \captionof{figure}{\textbf{a.} Memory signature of the first encoder layer of a Listen-Attend-Spell (LAS) model trained on WSJ. \textbf{b.} Memory signature of the last encoder layer for the same LAS model. \textbf{c.} Memory signature of the query vectors of the decoder layer for the same LAS model. In all figures, the the axis is the same as in Figure~\ref{fig:confusion}. Accuracy ranges from dark purple (100\% recall) to white (0\% recall).}\medskip

  \label{fig:las}
  \end{minipage}
\end{figure}

Much work in ASR in the past few years has been focused on developing end-to-end networks that read audio samples and and produce grapheme output \cite{wav2letter, speech2letters}. One such model that is commonly used for benchmarking is Listen-Attend-Spell (LAS) \cite{las}. The architecture in LAS has three or more stacked LSTM encoder layers, followed by one or more stacked LSTM decoder layers. At every time step $t$, the decoder cell receives (1) the previous decoder output and (2) a subset of encoder outputs, as decided by an attention filter produced by an attention network with input $q$, another decoder output labeled as the `attention query vector'. The decoder outputs a grapheme prediction at every time step. A common interpretation of LAS's audio encoder/language decoder architecture is that it functions as the acoustic/language models that compose traditional ASR systems. With this encoder-decoder architecture, however, some questions naturally arise: what sort of grapheme-level information is encoded in the recurrent state of the audio encoder? Is knowledge of prior input distributed evenly across encoder layers? Or does the decoder's recurrent state encode more information about prior graphemes?

We turn to memory signatures to provide us clues. Figure~\ref{fig:las} displays the character-level recall accuracies from states in two layers in the encoder, as well as from the pre-attention decoder output. Firstly, we notice the much stronger recall from the last encoder layer as compared to the first: the encoder appears accumulate knowledge of inputs from a wider time period as the layers grow deeper. This encoder behavior constrasts with the behavior noticed in recurrent language models from Figure~\ref{fig:statedepth}, where recall degraded deeper in the network. It is interesting that even using shallow networks, we are able to extract information about graphemes from the encoder layers, suggesting some grapheme symbolization is occurring in what is sometimes thought to analogous to the acoustic model.

The decoder output in LAS demonstrates a much stronger character recall than the encoder layers. This could be because either the decoder accumulates more information about the past, or that graphemes are easier to back-predict, because the decoder operates on graphemes, unlike the encoder. 

\section{Discussion}
\label{sec:conclusion}

In this paper, we describe a technique to visualize the recall capacity of LSTM/GRU kernels using decoders of recurrent state. Input recall is necessary to maintain on long-term dependencies, and we explicitly use this to characterize kernel memory. As demonstrated in our case studies, memory signatures are useful in gauging information diversity in recurrent kernels, visualizing information flow in end-to-end ASR networks, and understanding domain-specific language patterns. We encourage use of memory signatures, for example, to understand a language model's ability to track inter-word dependencies and retain knowledge of previously encountered tokens.

While our experiments test recall of grapheme input, it is useful to note that memory signatures are not limited to tracking recall of single sequence input; the same decoders could test recall of prior n-grams or discrete audio features. These is an avenue for future work.

\newpage


\vfill\pagebreak

\bibliographystyle{IEEEbib}
\bibliography{strings,refs}

\begin{thebibliography}{10}

\bibitem{lstmlm}
Martin Sundermeyer, Ralf Schl{\"u}ter, and Hermann Ney,
\newblock ``{LSTM} neural networks for language modeling,''
\newblock in {\em Thirteenth Annual Conference of the International Speech
  Communication Association}, 2012.

\bibitem{lstmam}
Ha{\c{s}}im Sak, Andrew Senior, and Fran{\c{c}}oise Beaufays,
\newblock ``Long short-term memory recurrent neural network architectures for
  large scale acoustic modeling,''
\newblock 2014.

\bibitem{las}
William Chan, Navdeep Jaitly, Quoc~V Le, and Oriol Vinyals,
\newblock ``{Listen, Attend and Spell},''
\newblock {\em arXiv preprint arXiv:1508.01211}, 2015.

\bibitem{translation}
Dzmitry Bahdanau, Kyunghyun Cho, and Yoshua Bengio,
\newblock ``Neural machine translation by jointly learning to align and
  translate,''
\newblock {\em arXiv preprint arXiv:1409.0473}, 2014.

\bibitem{retraining}
Deirdre Hogan, Jennifer Foster, Joachim Wagner, and Josef Van~Genabith,
\newblock ``Parser-based retraining for domain adaptation of probabilistic
  generators,''
\newblock in {\em Proceedings of the Fifth International Natural Language
  Generation Conference}. Association for Computational Linguistics, 2008, pp.
  165--168.

\bibitem{factorized}
Lahiru Samarakoon and Khe~Chai Sim,
\newblock ``Factorized hidden layer adaptation for deep neural network based
  acoustic modeling,''
\newblock {\em IEEE/ACM Transactions on Audio, Speech, and Language
  Processing}, vol. 24, no. 12, pp. 2241--2250, 2016.

\bibitem{stimulatedlearning}
Shawn Tan, Khe~Chai Sim, and Mark Gales,
\newblock ``Improving the interpretability of deep neural networks with
  stimulated learning,''
\newblock in {\em Automatic Speech Recognition and Understanding (ASRU), 2015
  IEEE Workshop on}. IEEE, 2015, pp. 617--623.

\bibitem{hmmrnn}
Viktoriya Krakovna and Finale Doshi-Velez,
\newblock ``Increasing the interpretability of recurrent neural networks using
  hidden {Markov} models,''
\newblock {\em arXiv preprint arXiv:1606.05320}, 2016.

\bibitem{affinenetworks}
Jakob~N Foerster, Justin Gilmer, Jascha Sohl-Dickstein, Jan Chorowski, and
  David Sussillo,
\newblock ``{Input Switched Affine Networks: An RNN Architecture Designed for
  Interpretability},''
\newblock in {\em International Conference on Machine Learning}, 2017, pp.
  1136--1145.

\bibitem{khechaiinterp2}
Khe~Chai Sim,
\newblock ``{Sensitivity-Characterised Activity Neurogram (SCAN) for
  Visualising and Understanding the Inner Workings of Deep Neural Network},''
\newblock {\em IEICE TRANSACTIONS on Information and Systems}, vol. 99, no. 10,
  pp. 2423--2430, 2016.

\bibitem{patternnet}
Pieter-Jan Kindermans, Kristof~T Sch{\"u}tt, Maximilian Alber, Klaus-Robert
  M{\"u}ller, and Sven D{\"a}hne,
\newblock ``{PatternNet and PatternLRP--Improving the interpretability of
  neural networks},''
\newblock {\em arXiv preprint arXiv:1705.05598}, 2017.

\bibitem{activationmaximization}
Karen Simonyan, Andrea Vedaldi, and Andrew Zisserman,
\newblock ``{Deep inside convolutional networks: Visualising image
  classification models and saliency maps},''
\newblock {\em arXiv preprint arXiv:1312.6034}, 2013.

\bibitem{deconvolution}
Matthew~D Zeiler and Rob Fergus,
\newblock ``Visualizing and understanding convolutional networks,''
\newblock in {\em European Conference on Computer Vision}. Springer, 2014, pp.
  818--833.

\bibitem{pertubingweights}
Julian~D Olden and Donald~A Jackson,
\newblock ``Illuminating the “black box”: a randomization approach for
  understanding variable contributions in artificial neural networks,''
\newblock {\em Ecological modelling}, vol. 154, no. 1, pp. 135--150, 2002.

\bibitem{deeplift}
Avanti Shrikumar, Peyton Greenside, Anna Shcherbina, and Anshul Kundaje,
\newblock ``{Not Just A Black Box: Interpretable Deep Learning by Propagating
  Activation Differences},''
\newblock ICML, 2016.

\bibitem{extra1}
Leila Arras, Gr{\'e}goire Montavon, Klaus-Robert M{\"u}ller, and Wojciech
  Samek,
\newblock ``Explaining recurrent neural network predictions in sentiment
  analysis,''
\newblock {\em arXiv preprint arXiv:1706.07206}, 2017.

\bibitem{extra2}
Sebastian Bach, Alexander Binder, Gr{\'e}goire Montavon, Frederick Klauschen,
  Klaus-Robert M{\"u}ller, and Wojciech Samek,
\newblock ``On pixel-wise explanations for non-linear classifier decisions by
  layer-wise relevance propagation,''
\newblock {\em PloS one}, vol. 10, no. 7, pp. e0130140, 2015.

\bibitem{icasspviz}
Gr{\'e}goire Montavon, Wojciech Samek, and Klaus-Robert M{\"u}ller,
\newblock ``Methods for interpreting and understanding deep neural networks,''
\newblock {\em arXiv preprint arXiv:1706.07979}, 2017.

\bibitem{inversion}
Aravindh Mahendran and Andrea Vedaldi,
\newblock ``Understanding deep image representations by inverting them,''
\newblock in {\em Proceedings of the IEEE conference on computer vision and
  pattern recognition}, 2015, pp. 5188--5196.

\bibitem{nlplstm}
Jiwei Li, Xinlei Chen, Eduard Hovy, and Dan Jurafsky,
\newblock ``Visualizing and understanding neural models in {NLP},''
\newblock {\em arXiv preprint arXiv:1506.01066}, 2015.

\bibitem{karpathy}
Andrej Karpathy, Justin Johnson, and Li~Fei-Fei,
\newblock ``Visualizing and understanding recurrent networks,''
\newblock {\em arXiv preprint arXiv:1506.02078}, 2015.

\bibitem{ruleextraction}
W~James Murdoch and Arthur Szlam,
\newblock ``{Automatic Rule Extraction from Long Short Term Memory Networks},''
\newblock {\em arXiv preprint arXiv:1702.02540}, 2017.

\bibitem{lstm}
Sepp Hochreiter and J{\"u}rgen Schmidhuber,
\newblock ``Long short-term memory,''
\newblock {\em Neural computation}, vol. 9, no. 8, pp. 1735--1780, 1997.

\bibitem{grus}
Junyoung Chung, Caglar Gulcehre, Kyunghyun Cho, and Yoshua Bengio,
\newblock ``Gated feedback recurrent neural networks,''
\newblock in {\em International Conference on Machine Learning}, 2015, pp.
  2067--2075.

\bibitem{wsj}
Douglas~B Paul and Janet~M Baker,
\newblock ``{The Design for the Wall Street Journal-based CSR corpus},''
\newblock in {\em Proceedings of the workshop on Speech and Natural Language}.
  Association for Computational Linguistics, 1992, pp. 357--362.

\bibitem{crossdomain}
Bo-June~Paul Hsu and James Glass,
\newblock ``N-gram weighting: reducing training data mismatch in cross-domain
  language model estimation,''
\newblock in {\em Proceedings of the Conference on Empirical Methods in Natural
  Language Processing}. Association for Computational Linguistics, 2008, pp.
  829--838.

\bibitem{ngramlstmcombo}
Graham Neubig and Chris Dyer,
\newblock ``Generalizing and hybridizing count-based and neural language
  models,''
\newblock {\em arXiv preprint arXiv:1606.00499}, 2016.

\bibitem{ngramlstmcombo2}
Ciprian Chelba, Mohammad Norouzi, and Samy Bengio,
\newblock ``{N-gram Language Modeling using Recurrent Neural Network
  Estimation},''
\newblock {\em arXiv preprint arXiv:1703.10724}, 2017.

\bibitem{factorized2}
Lahiru Samarakoon, Brian Mak, and Khe~Chai Sim,
\newblock ``{Learning Factorized Transforms for Unsupervised Adaptation of
  LSTM-RNN Acoustic Models},''
\newblock {\em Proc. Interspeech 2017}, pp. 744--748, 2017.

\bibitem{wav2letter}
Ronan Collobert, Christian Puhrsch, and Gabriel Synnaeve,
\newblock ``Wav2letter: an end-to-end convnet-based speech recognition
  system,''
\newblock {\em arXiv preprint arXiv:1609.03193}, 2016.

\bibitem{speech2letters}
Florian Eyben, Martin W{\"o}llmer, Bj{\"o}rn Schuller, and Alex Graves,
\newblock ``From speech to letters-using a novel neural network architecture
  for grapheme based asr,''
\newblock in {\em Automatic Speech Recognition \& Understanding, 2009. ASRU
  2009. IEEE Workshop on}. IEEE, 2009, pp. 376--380.

\end{thebibliography}

\end{document}